\documentclass{article} % For LaTeX2e
\usepackage{EgoExoNext,times}

\usepackage{amsmath,amsfonts,bm}

\def\eqref#1{equation~\ref{#1}}
\def\1{\bm{1}}

\DeclareMathAlphabet{\mathsfit}{\encodingdefault}{\sfdefault}{m}{sl}
\SetMathAlphabet{\mathsfit}{bold}{\encodingdefault}{\sfdefault}{bx}{n}

\usepackage{hyperref}
\usepackage{url}

\usepackage{algorithm}
\usepackage{algorithmic}

\usepackage{newfloat}
\usepackage{listings}
\usepackage{xcolor}
\usepackage{graphicx}

\usepackage{newfloat}
\usepackage{listings}
\usepackage{xcolor}

\usepackage{booktabs}

\usepackage[table]{xcolor}
\usepackage{multirow}
\usepackage[most]{tcolorbox}

\newtcolorbox{captionpromptbox}[2][]{%
enhanced,
sharp corners,
boxrule=0.4pt,
colback=gray!3!white,
colframe=gray!60!black,
colbacktitle=gray!6!white,
coltitle=black,
title={#2},
fonttitle=\bfseries,
borderline west={2pt}{0pt}{blue!35!black},
opacityback=0.96,
opacitybacktitle=0.92,
left=8pt,
right=8pt,
top=6pt,
bottom=6pt,
width=\linewidth,
#1
}

\title{EgoExo-Next:Benchmarking Vision-Language Models on Visual-Option Next-State and Cross-View Reasoning}

\author{
Yutong Li\textsuperscript{1},
Molin Wang\textsuperscript{2},
Xiaotong Li\textsuperscript{1},
Yanyan Fang\textsuperscript{1},
Daoguo Dong\textsuperscript{1},
Ziyi Ye\textsuperscript{1}
\\[6pt]
\textsuperscript{1}Fudan University, Shanghai, China
\\
\textsuperscript{2}Xi'an Jiaotong-Liverpool University, Suzhou, China
}

\EgoExoNextfinalcopy % Uncomment for camera-ready version, but NOT for submission.
\begin{document}

\maketitle

\begin{abstract}
Vision-language models (VLMs) are increasingly evaluated for egocentric and cross-view video reasoning, yet existing benchmarks largely focus on semantic event understanding, temporal relations, or correspondence between already observed views, leaving their ability to reason directly about future visual states underexplored. We introduce EgoExo-Next, a visual-option benchmark for dynamic visual-state reasoning, where models must identify how an observed action trajectory subsequently appears rather than predict only an action label or textual description. EgoExo-Next contains 2,503 human-curated four-choice questions from six public egocentric and ego--exo video sources and comprises four interconnected subtasks that evaluate egocentric next-state prediction, bidirectional ego--exo state correspondence, exocentric next-state prediction, and their composition in Ego-to-Exo Next-State. Extensive evaluation of proprietary, open-source, and spatial reasoning VLMs reveals a substantial human--model gap, with the best model achieving 43.81\% average accuracy compared with 98.55\% for humans, and the largest degradation occurring on the composed Ego-to-Exo task. These results suggest that current VLMs remain substantially limited in dynamic visual-state reasoning, particularly when temporal progression and cross-view reasoning must be composed. The benchmark is publicly available at \url{https://huggingface.co/datasets/yutongli2024/EgoExo-Next}.
\end{abstract}

\section{Introduction}

Vision-language models (VLMs) have demonstrated impressive reasoning capabilities across a wide range of multimodal tasks~\citep{VLMSurvey1, VLMSurvey2}. However, many existing evaluations still formulate reasoning through language-centered outputs, such as answering questions, describing events, or selecting among textual statements~\citep{SpatialReasoningSurvey1, SpatialReasoningSurvey2}. In physical environments, intelligent agents must also reason about how visual states evolve. Given an ongoing action, a model should be able to infer how the observed scene progresses over time, determine how the same state appears from another viewpoint, and combine these abilities to reason about a future state across perspectives~\citep{EmbodiedAI_survey1}.

Humans perform such reasoning naturally. From first-person observations of a hand reaching for a tool or manipulating an object, we can anticipate the subsequent object configuration and infer the corresponding body pose and scene layout from an external viewpoint~\citep{cognitive1, cognitive2}. This goes beyond recognizing what action is occurring: it requires reasoning about both \textit{temporal progression}, i.e., how the observable physical configuration changes over time, and \textit{viewpoint transformation}, i.e., how the same state is represented from another viewpoint. Their composition is particularly relevant to embodied intelligence, where an agent may perceive the environment egocentrically while demonstrations, supervision, or collaborative observations are available from exocentric views~\citep{EmbodiedAI_survey2}.

\begin{figure*}
    \centering
    \includegraphics[width=1\linewidth]{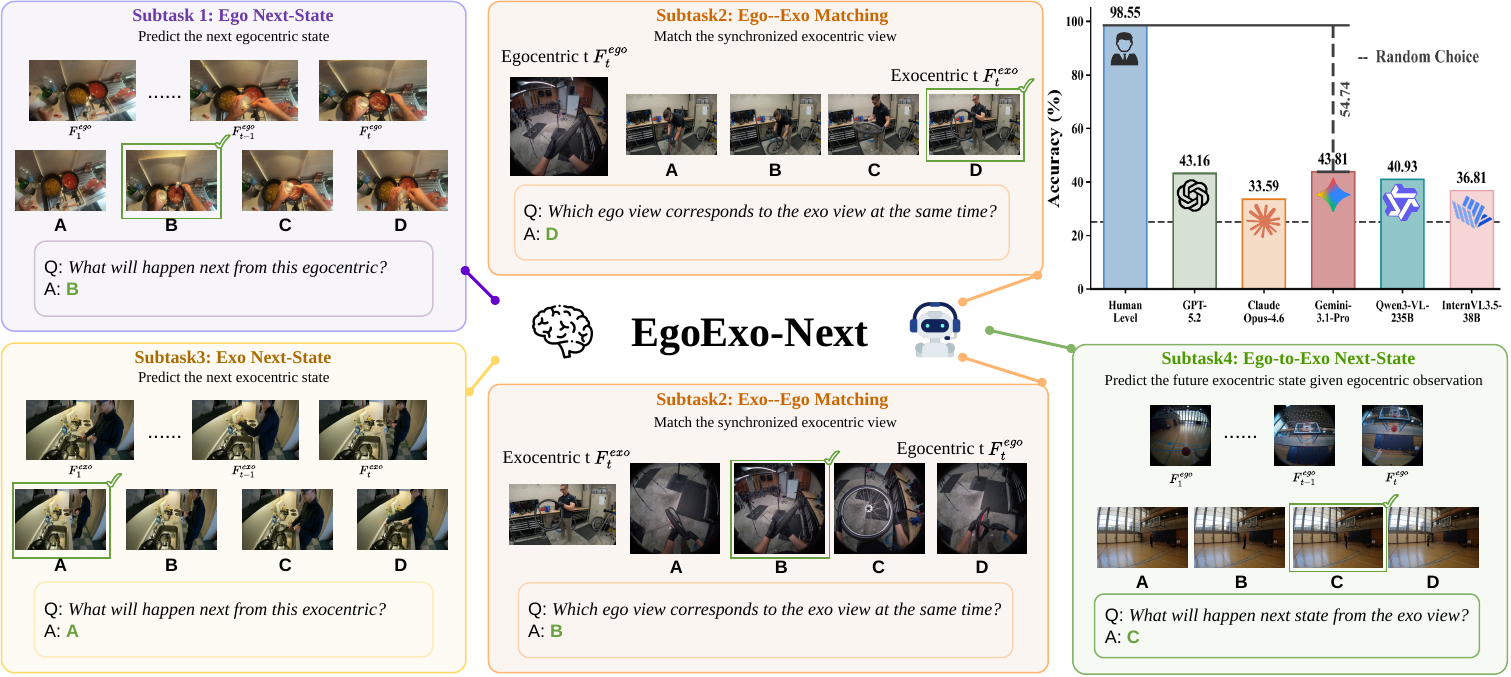}
    \caption{The EgoExo-Next benchmark. Main panels: Examples of four subtasks for next-state prediction and ego-exo view matching. Upper right: Comparison of model and human performance, showing a gap between current VLMs and humans.}
    \vspace{-0.5cm}
    \label{fig:Introduction}
\end{figure*}

Existing benchmarks have established important foundations for egocentric and ego--exo video reasoning. EgoSchema~\citep{egoschema} evaluates long-form temporal understanding in egocentric videos, EgoPlan~\citep{egoplan} focuses on goal-conditioned next-action planning, and EgoTempo~\citep{egotempo} probes fine-grained temporal understanding over egocentric video sequences. More closely related to our setting, EgoExoBench~\citep{egoexobench} evaluates cross-view understanding through semantic alignment, viewpoint association, and temporal reasoning, including action prediction, action ordering, sequence alignment, and cross-view matching. Despite this progress, these benchmarks mainly evaluate whether models can recognize, predict, order, or align actions and events, or establish correspondence between ego--exo observations that are already given. They do not directly test whether a model can identify how the underlying visual state itself progresses into an unobserved future state. 

However, reasoning about the future visual state requires a different level of understanding. Predicting what action happens next is not equivalent to identifying what the scene should look like next: an ongoing action may pass through multiple visually similar yet physically distinct states that differ in object configuration, body pose, contact relations, and spatial layout. A model may therefore correctly infer what happens next at the semantic level while still failing to identify what the scene looks like next. The distinction is even more pronounced in the ego--exo setting. Cross-view matching evaluates correspondence between already observed views, whereas EgoExo-Next requires a model to first infer an unobserved future state from preceding egocentric observations and then determine how that future state appears from an exocentric viewpoint. Thus, EgoExo-Next targets a distinct capability beyond conventional ego--exo correspondence or temporal QA: dynamic visual-state reasoning, where models identify subsequent states directly in the visual space and, in the composed setting, combine temporal progression with viewpoint transformation.

To evaluate this capability directly, we introduce \textbf{EgoExo-Next}, a visual-option benchmark in which the prediction target is the subsequent visual state itself. Rather than mapping visual evidence to an action label, textual description, or temporal relation, models must identify the target state from four temporally related image candidates. The next-state tasks further include both earlier-state and later-future distractors, so the correct answer cannot be determined simply by identifying the only future state or selecting the most temporally advanced candidate. This formulation requires models to distinguish fine-grained progression within an action trajectory, where candidate states may share the same high-level semantics but differ in object configuration, body pose, contact relations, and spatial layout. EgoExo-Next then decomposes this capability into temporal visual-state progression and cross-view state correspondence, and evaluates their composition in a more demanding setting. In particular, EgoExo-Next requires a model to infer an unobserved subsequent state from preceding egocentric observations and identify how that state appears from an exocentric viewpoint. This composition goes beyond either next-action prediction or matching already observed ego--exo content.

EgoExo-Next contains 2,503 human-curated four-choice questions constructed from six publicly available egocentric and ego--exo video sources, covering diverse indoor and outdoor scenarios. As illustrated in Fig.~\ref{fig:Introduction}, the benchmark comprises four interconnected subtasks. \textbf{Subtask 1: Ego Next-State} evaluates temporal visual-state progression from the egocentric perspective. \textbf{Subtask 2: Ego--Exo Matching} evaluates bidirectional correspondence between synchronized egocentric and exocentric states at the same time. \textbf{Subtask 3: Exo Next-State} evaluates temporal visual-state progression from the exocentric perspective. Finally, \textbf{Subtask 4: Ego-to-Exo Next-State} composes temporal progression with viewpoint transformation, requiring models to identify the future exocentric visual state implied by preceding egocentric observations.

We conduct extensive experiments on representative proprietary, open-source, and spatial reasoning models. Across all four subtasks, current VLMs remain substantially below human performance, with the best model achieving an average accuracy of 43.81\%, compared with 98.55\% for humans. The largest gap appears on \textbf{Ego-to-Exo Next-State}, where models must infer an unobserved subsequent state from preceding egocentric observations and identify how that state appears from an exocentric viewpoint. The pronounced degradation on this composed task highlights the difficulty of combining temporal visual-state progression with viewpoint transformation. Controlled analyses further show that performance depends on observation context, temporal order, and ego--exo correspondence rather than candidate-side cues alone. Together, these results establish EgoExo-Next as a challenging diagnostic benchmark for evaluating temporal visual-state progression, cross-view correspondence, and their composition.

In summary, our contributions are as follows:
\begin{itemize}

\item We introduce \textbf{EgoExo-Next}, a visual-option benchmark for \textbf{dynamic visual-state reasoning}, including a composed \textbf{Ego-to-Exo Next-State} task that jointly evaluates temporal progression and viewpoint transformation.

\item We construct 2,503 human-curated four-choice questions from six public egocentric and ego--exo video sources, with four interconnected subtasks and a shortcut-resistant design based on non-uniform temporal sampling and controlled earlier-state and later-future distractors.

\item We benchmark representative proprietary, open-source, and spatial reasoning VLMs, revealing substantial human--model gaps and showing that successful prediction depends on temporal context and cross-view evidence rather than candidate-side shortcuts alone.

\end{itemize}

\section{Related Work}

\subsection{Egocentric VQA Benchmarks}

Egocentric vision offers a natural perspective for studying human-centered perception. Recent egocentric VQA benchmarks evaluate multimodal reasoning over egocentric videos. EgoSchema targets long-form temporal understanding in egocentric streams~\citep{egoschema}, while EgoPlan studies goal-oriented planning from ongoing activities~\citep{egoplan}. EgoThink evaluates first-person reasoning across multiple perceptual and cognitive dimensions~\citep{egothink}. EgoTempo specifically targets fine-grained temporal understanding and is designed to require reasoning over temporal patterns rather than relying on sparse frames or commonsense priors~\citep{egotempo}. Earlier work such as EgoVQA also explored first-person video question answering, providing an initial testbed for egocentric VideoQA~\citep{egovqa}. More recently, EgoProx evaluates egocentric 3D proximity reasoning over body-object spatial relations, while EgoExoBench assesses cross-view video understanding through MCQs on semantic alignment, viewpoint association, and temporal reasoning between first- and exocentric videos~\citep{egoprox, egoexobench}.

\subsection{Ego-Exo Datasets}

Paired egocentric--exocentric videos are important for studying cross-view perception and reasoning. Early datasets such as Charades-Ego~\citep{actor2018} provide wearable and external views of daily activities, while Home Action Genome~\citep{home2021} further introduces synchronized multi-view recordings with hierarchical activity annotations. Later datasets increasingly emphasize fine-grained procedural and interaction understanding. LEMMA~\citep{lemma} captures multi-agent, multi-task activities from multiple viewpoints, H2O~\citep{h2o} provides synchronized RGB-D observations with detailed hand--object pose annotations, and Assembly101~\citep{assembly101} offers large-scale egocentric and static-camera recordings of procedural assembly activities. Recent large-scale resources further broaden the scope of ego--exo learning. Ego-Exo4D~\citep{ego-exo4d} provides synchronized first- and third-person recordings of diverse skilled activities with rich multimodal annotations, while EgoExoLearn~\citep{egoexolearn} focuses on asynchronous cross-view learning between exocentric demonstrations and egocentric executions. EgoExo-Fitness~\citep{egoexofitness} additionally targets full-body action understanding and cross-view skill assessment.

\begin{figure*}[ht!]
    \centering
    \includegraphics[width=\linewidth]{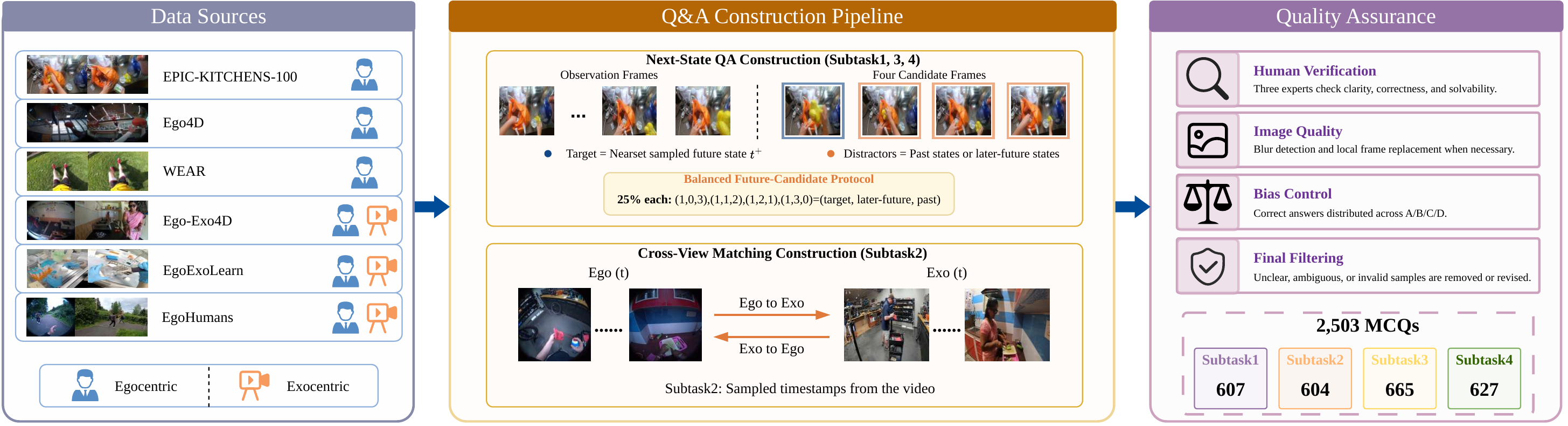}
    \includegraphics[width=\linewidth]{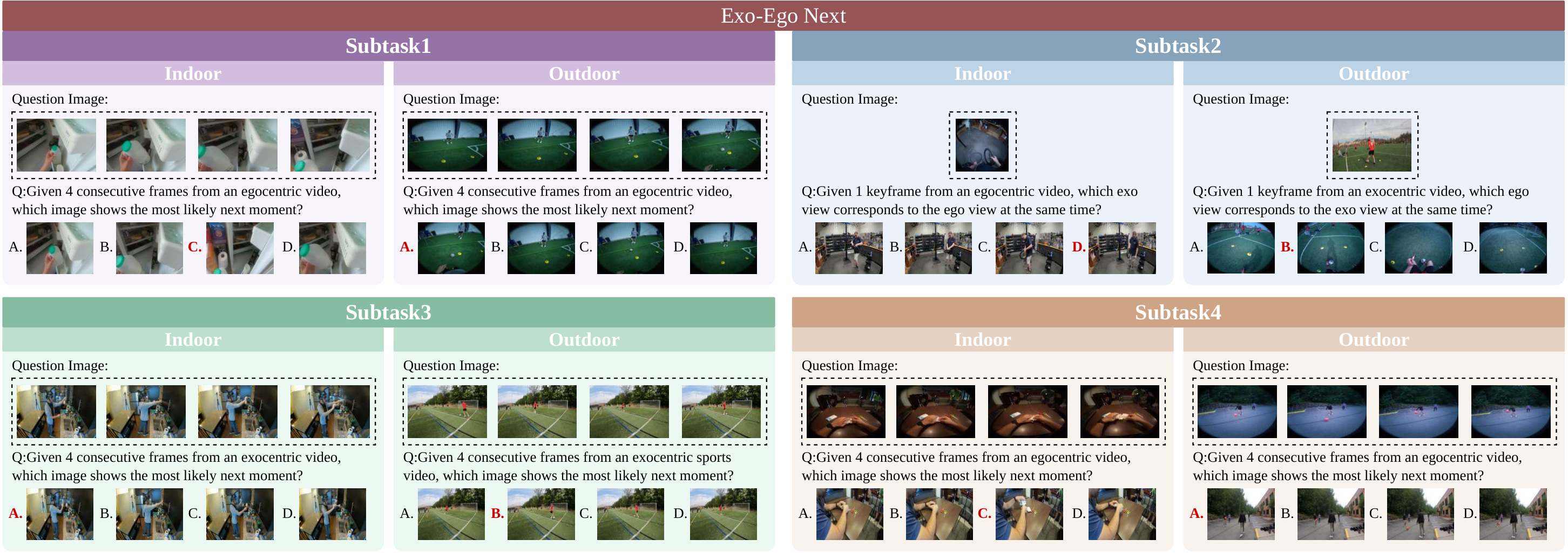}
    \caption{The upper part is the EgoExo-Next Pipeline. The lower part contains examples on EgoExo-Next. Correct answers are highlighted in red.}
    \vspace{-0.5cm}
    \label{fig:pipeline}
\end{figure*}

\section{EgoExo-Next}

In this section, we present EgoExo-Next, a visual-option benchmark for evaluating visual-state reasoning across egocentric and exocentric perspectives. We use \emph{visual state} to refer to the observable configuration of a scene captured in a video frame, including object configuration, human pose, contact relations, and spatial layout. The benchmark asks models to identify the appropriate target state from four visual candidates. Through four interconnected subtasks, EgoExo-Next evaluates temporal state progression, cross-view state correspondence, and their composition. We first define the four subtasks and then describe the data sources, question construction, benchmark statistics, and quality-control process.

\subsection{Overview}

Given egocentric or exocentric observations, EgoExo-Next requires a VLM to select the target visual state from four image candidates. Unlike benchmarks that mainly express temporal reasoning through action labels or textual descriptions, our next-state tasks directly evaluate whether models can identify how an observed trajectory subsequently appears in the visual state space.

Let $F_t^{ego}$ and $F_t^{exo}$ denote egocentric and exocentric visual states at time $t$, respectively, and let $F_{1:t}^{ego}$ and $F_{1:t}^{exo}$ denote temporally ordered observation sequences. We use $t^{+}>t$ to denote the target state following the final observation; the temporal interval between observations and the target is not fixed.

\textbf{Ego Next-State (Subtask 1)} evaluates $F_{1:t}^{ego}\rightarrow F_{t^{+}}^{ego}$, measuring temporal visual-state progression from the first-person perspective.
\textbf{Ego--Exo Matching (Subtask 2)} evaluates $F_t^{ego}\leftrightarrow F_t^{exo}$, measuring bidirectional correspondence between synchronized egocentric and exocentric states.
\textbf{Exo Next-State (Subtask 3)} evaluates $F_{1:t}^{exo}\rightarrow F_{t^{+}}^{exo}$, measuring temporal visual-state progression from the third-person perspective.
\textbf{Ego-to-Exo Next-State (Subtask 4)} evaluates $F_{1:t}^{ego}\rightarrow F_{t^{+}}^{exo}$, requiring models to infer the subsequent state from egocentric observations and identify how it appears from an exocentric viewpoint.

The four subtasks form an interconnected diagnostic suite. Subtasks 1 and 3 evaluate temporal visual-state progression from two viewpoints, while Subtask 2 evaluates cross-view state correspondence. Subtask 4 constitutes the composed setting, requiring temporal progression from preceding egocentric observations together with cross-view reasoning about the resulting future state.

A key property of the next-state subtasks is that candidates are not restricted to earlier states. Incorrect candidates may occur either before or after the final observation. We explicitly introduce later-future distractors so that neither identifying a candidate as ``future'' nor selecting the most temporally advanced candidate is sufficient to determine the answer. All candidates are sampled from the same action trajectory, preserving scene and action semantics while requiring fine-grained reasoning about trajectory progression.

% 1. 如何选取数据，
% 2. 如何构造数据
\subsection{Benchmark Construction}

As shown in Fig.~\ref{fig:pipeline}, EgoExo-Next is constructed from human-identified action segments through a visual-option QA generation pipeline. We first collect action clips from public egocentric and ego--exo video sources, and then construct observation sequences and candidate states according to each subtask. For the next-state subtasks, temporal samples are not restricted to a fixed interval, and candidate sets contain controlled combinations of earlier states and later-future distractors. We finally apply human verification and quality filtering to ensure that each target state is visually distinguishable from the provided candidates. We next describe the data sources, question-answer construction, benchmark statistics, and quality-control process.

\subsubsection{Data Source}\label{data_source}

EgoExo-Next collects images from six publicly available sources: EPIC-KITCHENS-100~\citep{ECCV_EPICKITCHENS, TPAMI_EPIC-KITCHENS}, Ego4D~\citep{ego4d}, WEAR~\citep{WEAR}, Ego-Exo4D~\citep{ego-exo4d}, EgoExoLearn~\citep{egoexolearn}, and EgoHumans~\citep{egohumans}. These sources span diverse indoor and outdoor environments and cover cooking, object manipulation, sports, procedural activities, and human motion. EPIC-KITCHENS-100, Ego4D, and WEAR mainly provide egocentric videos for evaluating egocentric temporal visual-state progression, whereas Ego-Exo4D, EgoExoLearn, and EgoHumans provide paired or multi-view egocentric--exocentric recordings for cross-view state correspondence and composed ego-to-exo next-state reasoning. Together, these sources support the evaluation of temporal progression within each viewpoint, correspondence across viewpoints, and their composition.

\begin{figure}
    \centering
    \includegraphics[width=\linewidth]{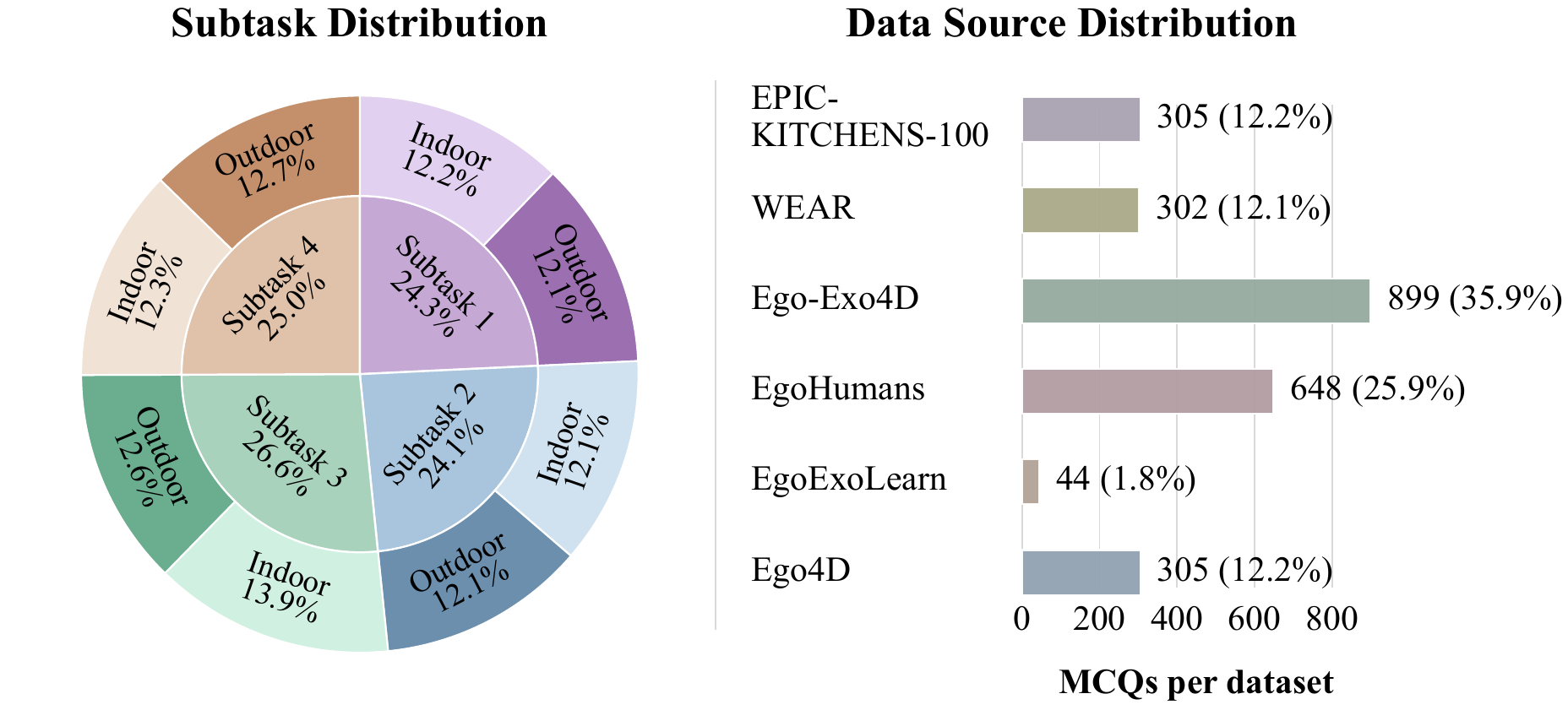}
    \caption{Benchmark composition by subtask, scene type, and data source.}
    \vspace{-0.5cm}
    \label{fig:benchmark_statistics}
\end{figure}

\subsubsection{Question-Answer Construction}

To construct the visual-option QA pairs, trained annotators first inspect the raw videos and identify complete action segments, which are then extracted using Python scripts. Since action durations and dynamics vary substantially, we do not impose a fixed temporal grid or constant sampling interval. Observation frames preserve their temporal order but may have different time gaps across and within clips. For Subtasks 1, 3, and 4, each question contains four candidate states sampled from the same action trajectory. Let $t$ denote the final observation time and $t^{+}>t$ the target time. The target is the nearest sampled future state following the observations, while distractors may occur either before $t$ or after $t^{+}$. We divide the questions equally into four groups according to the number of future candidates: 25\% contain one future candidate, 25\% contain two, 25\% contain three, and 25\% contain four. Accordingly, the four groups contain $(1,0,3)$, $(1,1,2)$, $(1,2,1)$, and $(1,3,0)$ target/later-future/past candidates, respectively. This design prevents the correct answer from being identified simply as the only future state or the most temporally advanced candidate.

\textbf{Subtask 1: Ego Next-State.}
A temporally ordered sequence of egocentric observations is provided, and the model selects $F_{t^{+}}^{ego}$ from four egocentric candidates constructed using the above past/future distractor protocol.

\textbf{Subtask 2: Ego--Exo Matching.}
For synchronized ego--exo videos, we sample matched timestamps and construct bidirectional ego-to-exo and exo-to-ego questions. Given a query frame from one viewpoint, the model selects its synchronized counterpart from four candidates in the other viewpoint.

\textbf{Subtask 3: Exo Next-State.}
Subtask 3 follows the same construction as Subtask 1 in the exocentric view, requiring the model to select $F_{t^{+}}^{exo}$ from temporally related exocentric candidates.

\textbf{Subtask 4: Ego-to-Exo Next-State.}
The model receives temporally ordered egocentric observations $F_{1:t}^{ego}$ and selects the corresponding future exocentric state $F_{t^{+}}^{exo}$. The exocentric candidates follow the same controlled past/future distractor construction, requiring both temporal progression and cross-view reasoning.

Across all subtasks, candidate order is randomized and correct-answer positions are balanced across A/B/C/D to reduce answer-position bias.

\subsubsection{Benchmark Statistics}

Following the construction pipeline in Fig.~\ref{fig:pipeline}, EgoExo-Next contains 2,503 four-way MCQs collected from six public video sources. As shown in Fig.~\ref{fig:benchmark_statistics}, Subtasks 1--4 account for 24.3\%, 24.1\%, 26.6\%, and 25.0\% of the benchmark, corresponding to 607, 604, 665, and 627 questions, respectively. Indoor and outdoor questions account for 50.5\% and 49.5\%, corresponding to 1,265 and 1,238 questions. For Subtasks 1, 3, and 4, questions are evenly distributed across groups containing one to four future candidates. EPIC-KITCHENS-100, WEAR, Ego-Exo4D, EgoHumans, EgoExoLearn, and Ego4D contribute 12.2\%, 12.1\%, 35.9\%, 25.9\%, 1.8\%, and 12.2\% of the benchmark, corresponding to 305, 302, 899, 648, 44, and 305 questions, respectively. The relatively small contribution of EgoExoLearn reflects our diversity-oriented curation: many candidate clips share similar scene and activity patterns with samples from other sources, so we retain 44 high-quality samples.

\subsubsection{Quality Assurance and Filtering}

To ensure benchmark quality and human-solvability, three additional reviewers, independent of the original annotators, systematically inspect every generated QA group using only the provided visual inputs. They flag samples with unclear actions, ambiguous trajectory progression, poor visual quality, indistinguishable candidate states, missing cross-view counterparts, or failed frame extraction. For Subtasks 1, 3, and 4, reviewers additionally verify that the target state is distinguishable from both earlier-state and later-future distractors based on the provided observations. A QA group is retained only when the target can be reliably identified; otherwise, it is revised or removed. We further apply blur detection and local frame replacement when necessary. Correct-answer positions are balanced across A/B/C/D, and candidate order is randomized to reduce multiple-choice bias.

\section{Evaluation on EgoExo-Next}

\subsection{Experiment Setup}

We evaluate EgoExo-Next in a zero-shot four-way MCQ setting, where observation frames and candidate images are provided as separate visual inputs and each model outputs one option label (A--D). No in-context examples or task-specific fine-tuning are used. We evaluate representative proprietary, open-source, and spatial reasoning VLMs. Each model is evaluated three times, and we report the mean accuracy. For Subtask 2, the two matching directions are evaluated independently and averaged as
$\mathrm{Acc}_{S2}=(\mathrm{Acc}_{E\rightarrow X}+\mathrm{Acc}_{X\rightarrow E})/2$. The random-guessing accuracy is 25\%. We additionally evaluate training-free CLIP- and DINO-based nearest-neighbor baselines on Subtasks 1, 3, and 4 to test whether next-state selection can be explained by generic visual similarity. Detailed model configurations, deployment settings, similarity-baseline implementation, and evaluation prompts are provided in the \textbf{Appendix}.

% --------Table1--------
\begin{table*}[t]
\centering
\caption{Performance on EgoExo-Next. Results are accuracy (\%). Avg. denotes the mean over all eight settings. Due to the substantially higher evaluation cost of GPT-6, we evaluate it only on Subtask 4 and therefore do not report results for Subtasks 1--3 or an overall average. \textbf{Bold} and underlined values indicate the best and second-best model results, respectively, excluding human performance.
}
\resizebox{\textwidth}{!}{
\begin{tabular}{lccccccccc}
\toprule
\multirow{2}{*}{Models} 
& \multicolumn{2}{c}{Subtask 1}
& \multicolumn{2}{c}{Subtask 2}
& \multicolumn{2}{c}{Subtask 3}
& \multicolumn{2}{c}{Subtask 4}
& \multirow{2}{*}{Avg.} \\
\cmidrule(lr){2-3}
\cmidrule(lr){4-5}
\cmidrule(lr){6-7}
\cmidrule(lr){8-9}

& Indoor & Outdoor
& Indoor & Outdoor
& Indoor & Outdoor
& Indoor & Outdoor
& \\

\midrule
Human Level
& 98.36 & 99.14
& 99.34 & 99.67
& 98.85 & 99.05
& 97.09 & 96.86
& 98.55 \\

\midrule
\rowcolor{gray!15}
\multicolumn{10}{l}{\textit{Proprietary Models}} \\

GPT-6
& N/A & N/A
& N/A & N/A
& N/A & N/A
& \textbf{62.23} & \textbf{58.14}
& N/A \\

GPT-5.2
& \underline{51.48} & \textbf{58.57}
& 46.53 & 29.31
& \textbf{48.14} & \underline{47.47}
& \underline{32.89} & 30.91
& \underline{43.16} \\

Claude-Opus-4.6
& 34.75 & 35.71
& 42.39 & 32.12
& 29.23 & 41.14
& 26.33 & 27.04
& 33.59 \\

Gemini-3.1-Pro
& \textbf{54.75} & 52.86
& \textbf{59.44} & \textbf{42.38}
& \underline{44.13} & 42.41
& 25.90 & 28.62
& \textbf{43.81} \\

\midrule
\rowcolor{gray!15}
\multicolumn{10}{l}{\textit{Open-Source Models}} \\

Qwen3-VL-235B
& 51.15 & \underline{58.29}
& \underline{50.00} & 26.82
& 38.40 & \textbf{47.78}
& 26.71 & 28.30
& 40.93 \\

Qwen3-VL-32B
& 46.89 & 43.71
& 47.02 & 26.66
& 37.82 & 34.60
& 25.25 & \underline{34.28}
& 37.03 \\

Qwen3-VL-8B
& 45.18 & 44.65
& 41.44 & 25.59
& 40.17 & 35.24
& 26.75 & 31.13
& 36.27 \\

InternVL3.5-8B
& 41.31 & 38.54
& 41.06 & 26.16
& 32.38 & 31.96
& 25.57 & 26.42
& 32.93 \\

InternVL3.5-38B
& 45.90 & 44.27
& 46.69 & 28.15
& 36.96 & 37.34
& 25.89 & 29.25
& 36.81 \\

LLaVA-OneVision-1.5-8B
& 39.02 & 37.58
& 39.74 & 25.50
& 31.23 & 30.70
& 25.24 & 25.79
& 31.85 \\

\midrule
\rowcolor{gray!15}
\multicolumn{10}{l}{\textit{Spatial Reasoning Models}} \\

SpatialVLM
& 38.36 & 36.31
& 41.39 & 28.15
& 33.24 & 32.91
& 26.21 & 27.04
& 32.95 \\

SpatialMLLM
& 41.97 & 40.76
& 44.70 & 30.13
& 35.53 & 36.08
& 26.86 & 28.62
& 35.58 \\

RoboBrain
& 46.23 & 44.90
& 48.34 & \underline{32.45}
& 38.68 & 39.24
& 28.16 & 30.50
& 38.56 \\

\bottomrule
\end{tabular}
}

\label{tab:benchmark_results}
\end{table*}

\subsection{Main Results}

Table~\ref{tab:benchmark_results} yields three main observations.

(1) A substantial human--model gap remains in dynamic visual-state reasoning. Among the models evaluated across all four subtasks, Gemini-3.1-Pro achieves the highest overall accuracy of 43.81\%, followed by GPT-5.2 at 43.16\%, while Qwen3-VL-235B is the strongest open-source model at 40.93\%. These results remain far below the human accuracy of 98.55\%, revealing substantial limitations in current VLMs' dynamic visual-state reasoning across temporal progression, cross-view correspondence, and their composition. Since GPT-6 is evaluated only on Subtask 4 due to evaluation cost, it is excluded from the overall-average comparison.  
(2) Ego-to-Exo Next-State remains a challenging but discriminative composed reasoning task. Among models evaluated across all four subtasks, Subtask 4 exhibits the strongest performance degradation, with the best Indoor and Outdoor accuracies reaching only 32.89\% and 34.28\%, respectively. In contrast, GPT-6 substantially improves Subtask 4 performance to 62.23\% on Indoor scenes and 58.14\% on Outdoor scenes, outperforming the previous best results by 29.34 and 23.86 percentage points. This substantial improvement indicates that stronger VLMs can make meaningful progress on the composed task, suggesting that EgoExo-Next is sensitive to advances in model capability rather than merely producing floor-level performance. Nevertheless, GPT-6 remains well below human performance of 97.09\% and 96.86\%, leaving gaps of 34.86 and 38.72 percentage points on Indoor and Outdoor scenes, respectively. These results show that inferring an unobserved subsequent state from egocentric observations and identifying how that state appears from an exocentric viewpoint remains challenging even for substantially stronger models. 
(3) Neither scaling nor spatial specialization provides a consistent solution across the full benchmark. Proprietary models generally achieve stronger performance, although Qwen3-VL-235B remains competitive and achieves the best result on Subtask 3 Outdoor among models evaluated across all four subtasks. Scaling improves performance within some model families, such as InternVL3.5, but the trend is not consistently monotonic across Qwen3-VL variants. Specialized spatial reasoning models likewise do not consistently outperform general-purpose VLMs. Together with the substantial improvement of GPT-6 on Subtask 4, these results suggest that increased model capability can substantially improve composed visual-state reasoning, while dynamic visual-state reasoning across the broader set of temporal and cross-view settings remains far from human-level performance.

% --------Table2--------
\begin{table*}[t]
\centering
\caption{Benchmark validity and shortcut analysis. Results are accuracy (\%) averaged over subtasks and scene splits.}
\resizebox{\textwidth}{!}{
\begin{tabular}{lcccccc}
\toprule
\textbf{Evaluation Setting}
& \textbf{GPT-5.2}
& \textbf{Gemini-3.1-Pro}
& \textbf{Qwen3-VL-235B}
& \textbf{Qwen3-VL-32B}
& \textbf{RoboBrain}
& \textbf{Avg.} \\
\midrule

\rowcolor{gray!15}
\multicolumn{7}{l}{\textit{Overall Shortcut Test: Subtasks 1, 2, 3, 4}} \\

Full Input
& 43.16
& 43.81
& 40.93
& 37.03
& 38.56
& \textbf{40.70} \\

Candidate Only
& 27.25
& 26.75
& 26.00
& 25.80
& 25.75
& 26.31 \\

\midrule
\rowcolor{gray!15}
\multicolumn{7}{l}{\textit{Temporal Dependency: Subtasks 1, 3, 4}} \\

Full Sequence
& 44.91
& 41.45
& 41.77
& 37.09
& 37.95
& \textbf{40.63} \\

Last Frame Only
& 38.20
& 36.80
& 35.90
& 33.40
& 33.50
& 35.56 \\

Shuffled Sequence
& 34.70
& 32.60
& 31.80
& 29.60
& 29.80
& 31.70 \\

\midrule
\rowcolor{gray!15}
\multicolumn{7}{l}{\textit{Cross-View Dependency: Subtasks 2, 4}} \\

Full Observation
& 34.91
& 39.09
& 32.96
& 33.30
& 34.86
& \textbf{35.02} \\

Mismatched Observation
& 26.40
& 27.20
& 25.80
& 25.90
& 26.30
& 26.32 \\

\bottomrule
\end{tabular}
}
\label{tab:shortcut_analysis}
\end{table*}

\subsection{Benchmark Validity and Shortcut Analysis}

We conduct controlled ablations on five representative models to test whether performance depends on the intended temporal and cross-view information rather than candidate-side cues alone. The full-input reference rows average all four subtasks (\textit{Full Input}), temporal Subtasks 1, 3, and 4 (\textit{Full Sequence}), or cross-view Subtasks 2 and 4 (\textit{Full Observation}). We compare them with three controls:
(1) Candidate-Side Cues: Candidate Only removes all observation images;
(2) Temporal Dependency: Last Frame Only retains only the final observation, while \textit{Shuffled Sequence} permutes the observation order; and
(3) Cross-View Dependency: Mismatched Observation replaces the query with another sample from the same data source and scene split.

As shown in Table~\ref{tab:shortcut_analysis}, \textit{Candidate Only} reduces average accuracy from 40.70\% to 26.31\%, close to random guessing. Last Frame Only and \textit{Shuffled Sequence} reduce temporal accuracy from 40.63\% to 35.56\% and 31.70\%, respectively, while Mismatched Observation reduces cross-view accuracy from 35.02\% to 26.32\%. These controls do not exclude every possible low-level cue, but they show that performance depends on observation context, temporal order, and ego--exo matching rather than candidate-side information alone.

% --------Table3--------
\begin{table*}[t]
\centering
\caption{
Training-free visual-similarity baselines on the three next-state subtasks. The left block reports accuracy by subtask, while the right block groups questions by the number of future candidates among the four options.
}
\resizebox{\textwidth}{!}{
\begin{tabular}{lcccc|ccccc}
\toprule
& \multicolumn{4}{c|}{\textbf{The Type of Subtask}}
& \multicolumn{5}{c}{\textbf{The Number of Future Candidates}} \\
\cmidrule(lr){2-5}
\cmidrule(lr){6-10}

\textbf{Method}
& \textbf{Subtask 1}
& \textbf{Subtask 3}
& \textbf{Subtask 4}
& \textbf{Avg.}
& \textbf{1}
& \textbf{2}
& \textbf{3}
& \textbf{4}
& \textbf{Avg.} \\
\midrule

Random
& 25.00
& 25.00
& 25.00
& 25.00
& 25.00
& 25.00
& 25.00
& 25.00
& 25.00 \\

CLIP
& 27.35
& 26.62
& 25.68
& 26.54
& 29.05
& 27.16
& 25.68
& 24.26
& 26.54 \\

DINO
& 27.02
& 27.22
& 25.20
& 26.49
& 28.42
& 27.79
& 26.11
& 23.63
& 26.49 \\

\bottomrule
\end{tabular}
}
\label{tab:similarity_baseline}
\end{table*}

\subsection{Visual-Similarity Baselines}

To examine whether the next-state tasks can be solved by generic visual similarity rather than trajectory reasoning, we evaluate two training-free nearest-neighbor baselines based on CLIP and DINO. For Subtasks 1, 3, and 4, each baseline independently encodes the final observation frame and the four candidate images, computes their cosine similarities, and selects the candidate with the highest similarity. As shown in Table~\ref{tab:similarity_baseline}, CLIP and DINO achieve average accuracies of 26.54\% and 26.49\%, respectively, only slightly above the 25\% random baseline. Their performance also varies across subtasks: CLIP performs slightly better on Subtasks 1 and 4, whereas DINO performs better on Subtask 3, suggesting that neither representation provides a consistently stronger similarity shortcut. We further group the questions by the number of future candidates. With only one future candidate, CLIP and DINO achieve 29.05\% and 28.42\%, respectively, indicating that visual similarity provides only limited cues even in the least ambiguous setting. Their performance gradually approaches chance as additional later-future distractors are introduced. With four future candidates, where every option occurs after the final observation, CLIP and DINO achieve 24.26\% and 23.63\%, respectively. These results indicate that generic nearest-frame similarity is insufficient for reliably identifying the target successor state. In particular, the limited advantage observed with a single future candidate disappears when multiple temporally plausible future states are included, supporting the effectiveness of the future-distractor construction in reducing similarity-based shortcuts.

\begin{figure*}[ht!]
    \centering
    \includegraphics[width=1\linewidth]{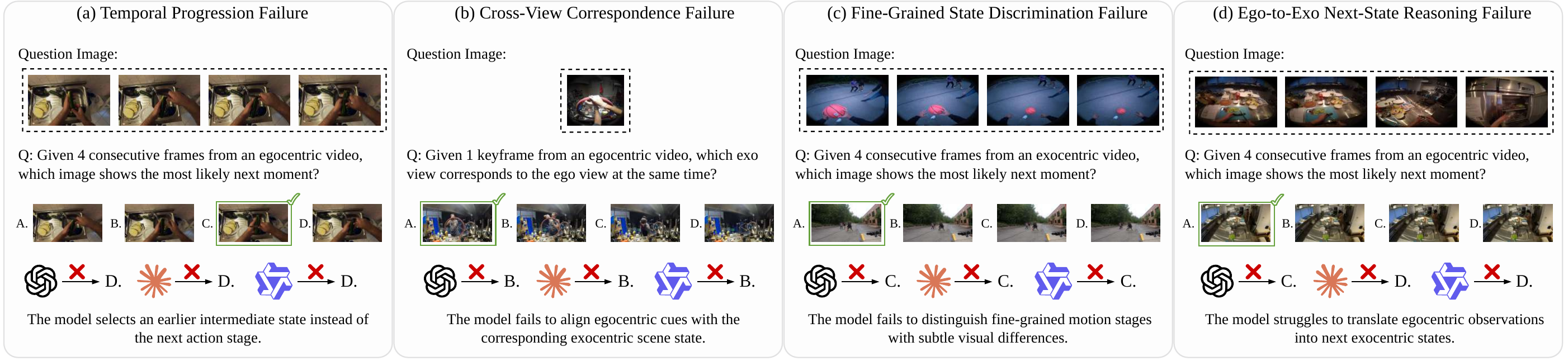}
    \vspace{-0.5cm}
    \caption{Representative failure cases on EgoExo-Next.}
    \label{fig:Failure_Case}
\end{figure*}

\subsection{Failure Case Analysis}

To further localize the limitations of current VLMs, we analyze representative failures from GPT-5.2, Claude-Opus-4.6, and Qwen3-VL-235B, covering both proprietary and open-source models. As shown in Fig.~\ref{fig:Failure_Case}, we identify four recurring failure types:
(1) Temporal Progression Failure: The model selects an earlier intermediate state rather than the subsequent action stage;
(2) Cross-View Correspondence Failure: The model fails to align local egocentric cues, such as hand--object interactions, with the corresponding exocentric scene configuration;
(3) Fine-Grained State Discrimination Failure: The model confuses visually similar candidates that differ only in subtle motion or state changes;
(4) Ego-to-Exo Next-State Reasoning Failure: The model fails to combine temporal progression and viewpoint transformation to identify the next exocentric state.
These cases reflect the diagnostic abilities evaluated by EgoExo-Next and show that current errors arise not only from action recognition, but also from temporal progression, cross-view correspondence, fine-grained state discrimination, and their composition.

\section{Conclusion}

We introduce EgoExo-Next, a visual-option benchmark for evaluating dynamic visual-state reasoning across temporal progression, cross-view correspondence, and their composition. It contains 2,503 human-curated four-way MCQs constructed from six public egocentric and ego--exo video sources and organized into four interconnected diagnostic subtasks. Extensive evaluations show that current proprietary, open-source, and spatial reasoning VLMs remain substantially below human performance, with the largest gap on the composed \textbf{Ego-to-Exo Next-State} task. Controlled analyses further show that performance depends on trajectory context, temporal organization, and ego--exo correspondence rather than candidate-side shortcuts alone. We hope EgoExo-Next supports the development of VLMs with stronger dynamic visual-state reasoning across time and perspective for embodied intelligence.

\bibliography{EgoExoNext}
\bibliographystyle{EgoExoNext}

\appendix
\newpage
\section*{Appendix}

\section{Appendix Overview and Organization}

This appendix provides supplementary details, benchmark comparisons, and diagnostic analyses to support and extend the main paper. The organization of the appendix is as follows:

\begin{enumerate}

    \item \textbf{Experimental Setup (Section~\ref{sec:experimental_setup}):}
    This section provides the complete experimental protocol for EgoExo-Next, including the zero-shot four-way multiple-choice setting, the proprietary, open-source, and spatial reasoning models evaluated, the aggregation strategy for bidirectional Ego--Exo Matching, and the training-free CLIP- and DINO-based visual-similarity baselines.

    \item \textbf{Evaluation Prompts (Section~\ref{sec:evaluation}):}
    We provide the complete evaluation prompts used for all four subtasks. Specifically, we present the unified next-state prompt for Ego Next-State, Exo Next-State, and Ego-to-Exo Next-State, as well as the bidirectional cross-view matching prompt for Ego--Exo Matching. We further describe how the prompt templates are instantiated according to observation perspective, candidate perspective, and scene setting.

    \item \textbf{Comparison with Existing Egocentric and Spatial Reasoning Benchmarks (Section~\ref{sec:benchmark_comparison}):}
    This section compares EgoExo-Next with representative egocentric, ego--exo, and spatial reasoning benchmarks, including EgoSchema, EgoPlan-Bench, EgoThink, VidEgoThink, EgoTempo, EgoProx, and EgoExoBench. We clarify that the distinction of EgoExo-Next lies not simply in temporal or cross-view reasoning individually, but in directly identifying an unobserved subsequent visual state and, in Ego-to-Exo Next-State, composing temporal visual-state progression with viewpoint transformation within a single prediction task.

    \item \textbf{Diagnosing Composed Ego-to-Exo Next-State Reasoning (Section~\ref{sec:composed_diagnosis}):}
    We further analyze the specific capabilities underlying the composed Ego-to-Exo Next-State task using a matched subset and controlled oracle inputs. By separately providing the current exocentric observation, the next-state egocentric observation, or both, we examine the respective contributions of cross-view correspondence, temporal visual-state progression, and their composition to the overall task difficulty.

    \item \textbf{Effect of Caption Augmentation (Section~\ref{sec:caption_augmentation}):}
    This section investigates whether explicit textual descriptions provide complementary semantic information for visual-state reasoning. We compare image-only evaluation with image-plus-caption evaluation using human-verified captions generated for candidate images, and analyze how explicit semantic cues affect model performance across different subtasks and scene settings.

    \item \textbf{Additional Implementation Details and Results (Section~\ref{sec:additional_details}):}
    Finally, we provide additional implementation and evaluation details, including model deployment and inference configurations as well as participant-level human evaluation results. These details supplement the main experiments and improve the reproducibility and transparency of our evaluation.

\end{enumerate}

Together, these supplementary materials provide additional methodological details, clarify the positioning of EgoExo-Next relative to existing benchmarks, and offer further evidence for the challenges of dynamic visual-state reasoning across temporal progression, cross-view correspondence, and their composition.

\section{Experimental Setup}\label{sec:experimental_setup}

EgoExo-Next is evaluated in a zero-shot four-way MCQ setting. Observation frames and candidate images are provided as separate visual inputs following Fig.~\ref{fig:pipeline}, and each model returns one option label (A--D). No in-context examples, task-specific fine-tuning, or gradient updates are used. All models receive semantically consistent task instructions and identical output constraints.

We evaluate three model groups:
(1) \textit{Proprietary Models}, including GPT-6~\citep{gpt6astra}, GPT-5.2~\citep{singh2025openai}, Claude-Opus-4.6~\citep{anthropic_claude4_2025}, and Gemini-3.1-Pro~\citep{team2023gemini};
(2) \textit{Open-source Models}, including Qwen3-VL~\citep{bai2025qwen3}, InternVL3.5~\citep{zhu2025internvl3}, and LLaVA-OneVision~\citep{llava-onevision};
and (3) \textit{Spatial Reasoning Models}, including SpatialVLM~\citep{chen2024spatialvlm}\footnote{\url{https://github.com/remyxai/VQASynth}}, SpatialMLLM~\citep{wu2026spatialmllm}\footnote{\url{https://github.com/THU-SI/Spatial-MLLM}}, and RoboBrain~\citep{ji2025robobrain}\footnote{\url{https://github.com/FlagOpen/RoboBrain}}.

The proprietary models and Qwen3-VL-235B are accessed through their official APIs, while the remaining models are deployed locally using their released implementations or checkpoints. Each model is evaluated three times, and the mean accuracy is reported. For Subtask 2, ego-to-exo and exo-to-ego matching are evaluated independently, and their mean accuracy is computed as $ \mathrm{Acc}_{S2} = \frac{ \mathrm{Acc}_{E\rightarrow X} + \mathrm{Acc}_{X\rightarrow E} }{2}.$ The random-guessing accuracy for all four-way questions is 25\%.

To test whether the next-state tasks can be solved through low-level visual similarity, we additionally evaluate training-free CLIP- and DINO-based nearest-neighbor baselines on Subtasks 1, 3, and 4. Each baseline independently encodes the final observation frame and the four candidate images, computes cosine similarity between the observation representation and each candidate representation, and selects the candidate with the highest similarity. These baselines use neither the preceding observation sequence nor task-specific training, providing a direct test of whether generic visual similarity alone is sufficient for next-state selection.

The complete evaluation prompts for all four subtasks, together with additional implementation and inference details, are provided in the following sections of the Appendix~\ref{sec:evaluation}.

\section{Evaluation Prompts}\label{sec:evaluation}

\subsection{Evaluation Prompts for Subtask1, 3, 4}\label{sec:subtask134_evaluation_prompts}

For Subtasks 1, 3, and 4, we adopt a unified next-state evaluation prompt. The prompt is instantiated according to the observation perspective and the perspective of the candidate images. Indoor and Outdoor settings differ only in the scene description and the visual cues emphasized in the instruction.

For Subtask 1, \texttt{<Observation Source>} is instantiated as ``an egocentric action video,'' while \texttt{<Observation Perspective>} is ``egocentric.'' The candidate image type is simply ``image,'' and \texttt{<Candidate View Description>} is instantiated as ``from the same egocentric perspective.'' The cross-view instruction is omitted.

For Subtask 3, the corresponding values are ``an exocentric action video,'' ``exocentric,'' ``image,'' and ``from the same exocentric perspective,'' respectively. The cross-view instruction is likewise omitted.

For Subtask 4, \texttt{<Observation Source>} is ``an egocentric video,'' \texttt{<Observation Perspective>} is ``egocentric,'' and \texttt{<Candidate Image Type>} is ``exocentric-view image.'' The candidate view description is instantiated as ``from an exocentric perspective,'' followed by the additional instruction: ``The exocentric view captures the same action from an external camera angle.''

\subsection{Evaluation Prompt for Subtask 2}\label{sec:subtask2_evaluation_prompts}

Subtask 2 evaluates cross-view correspondence between synchronized egocentric and exocentric observations. Each question contains one query keyframe from one perspective and four candidate images from the other perspective, and the model must identify the candidate depicting the same moment as the query. We use a unified cross-view matching prompt template for both the ego-to-exo and exo-to-ego directions.

For the ego-to-exo direction, \texttt{<Query View>} is instantiated as ``egocentric,'' and \texttt{<Candidate View>} as ``exocentric.'' The model is therefore asked to identify which exocentric candidate corresponds to the egocentric query image at the same time. For the exo-to-ego direction, \texttt{<Query View>} is instantiated as ``exocentric,'' and \texttt{<Candidate View>} as ``egocentric,'' requiring the model to identify the synchronized egocentric counterpart of the exocentric query.

For Indoor and Outdoor scenes, \texttt{<Setting>} is instantiated as ``indoor'' and ``outdoor,'' respectively. In both directions, the model is instructed to compare cross-view evidence including the visible action, human pose, object configuration, spatial layout, and scene context to identify the candidate depicting the same moment from the other viewpoint.

For each direction, the query keyframe is presented before the four candidate images. In the ego-to-exo direction, the query is labeled as \texttt{Ego-centric keyframe 1:}, followed by \texttt{Exo-centric option A:} to \texttt{Exo-centric option D:}. In the exo-to-ego direction, the query is labeled as \texttt{Exo-centric keyframe 1:}, followed by \texttt{Ego-centric option A:} to \texttt{Ego-centric option D:}. The two directions are evaluated independently, and their mean accuracy is reported as the Subtask 2 result.

%---------------Template for Subtask 1, 3, 4

\begin{captionpromptbox}{Template for Subtask 1, 3 and 4}
\textbf{Question:}

Given 4 temporally ordered observation frames from 
\texttt{<Observation Source>}, which 
\texttt{<Candidate Image Type>} shows the subsequent visual state following the observation sequence?

The four images above are temporally ordered observations from an 
\texttt{<Observation Perspective>} camera in a 
\texttt{<Setting>} environment. Options A--D show candidate visual states sampled from the same action trajectory 
\texttt{<Candidate View Description>}.

\texttt{<Cross-View Instruction>}

Observe how the action progresses across the observation frames---track 
\texttt{<Visual Cues>}---then select the candidate that depicts the subsequent state following the observation sequence.

Only output one letter: A, B, C, or D.
\end{captionpromptbox}

%---------------Template for Subtask 2

\begin{figure}[t]
\centering
\begin{captionpromptbox}{Template for Subtask 2}
\textbf{Question:}

Which \texttt{<Candidate View>} image corresponds to the \texttt{<Query View>} image at the same time?

The image above is a keyframe from the \texttt{<Query View>} perspective in a \texttt{<Setting>} environment. Options A--D each show a candidate image from the \texttt{<Candidate View>} perspective.

Observe the query frame carefully and compare the visible action, human pose, object configuration, spatial layout, and other cross-view cues. Select the candidate image that depicts the same moment from the corresponding \texttt{<Candidate View>} perspective.

Only output one letter: A, B, C, or D.
\end{captionpromptbox}

\caption{Unified cross-view matching prompt template for the two directions of Subtask 2. The placeholders are instantiated according to the query and candidate viewpoints.}
\label{fig:subtask2_prompt}
\end{figure}

\section{Comparison with Existing Egocentric and Spatial Reasoning Benchmarks}
\label{sec:benchmark_comparison}

\begin{table*}[t]
\caption{ Comparison with representative egocentric, ego--exo, and spatial reasoning benchmarks. ``Visual State'' indicates whether the prediction target itself is an unobserved subsequent visual state rather than an action, event, relation, or textual description.}
\label{tab:benchmark_comparison}
\centering
\resizebox{\textwidth}{!}{
\begin{tabular}{l l l l l c c@{}}
\toprule
\textbf{Benchmark}
& \textbf{Perspective}
& \textbf{Input Type}
& \textbf{Primary Evaluation Target}
& \textbf{Answer Type}
& \textbf{Cross-view}
& \shortstack{\textbf{Visual State}} \\
\midrule

EgoSchema
& Ego
& Video
& Long-form temporal understanding
& Text
& N/A
& N/A \\

EgoPlan-Bench
& Ego
& Video\& Image
& Goal-conditioned next-action planning
& Text
& N/A
& N/A \\

EgoThink
& Ego
& Image
& First-person perception\&reasoning
& Text
& N/A
& N/A \\

VidEgoThink
& Ego
& Video
& VQA\& planning\& grounding\& reward
& Text 
& N/A
& N/A \\

EgoTempo
& Ego
& Video
& Fine-grained temporal understanding
& Text
& N/A
& N/A \\

EgoProx
& Ego
& Video\& Frames
& 3D proximity\&spatial reasoning
& Text
& N/A
& N/A \\

EgoExoBench
& Ego--Exo
& Paired Video
& Semantic\&viewpoint\&temporal reasoning
& Image\&Text  
& Yes
& N/A \\

\midrule

\textbf{EgoExo-Next}
& \textbf{Ego\& Ego--Exo}
& \textbf{Multi-Image}
& \textbf{Visual-state progression\&correspondence}
& \textbf{Image}
& \textbf{Yes}
& \textbf{Yes} \\

\bottomrule
\end{tabular}
}
\end{table*}

Existing benchmarks cover several important dimensions of egocentric and embodied reasoning, but differ substantially in their prediction targets. EgoSchema~\citep{egoschema} evaluates long-form temporal understanding, while EgoPlan-Bench~\citep{egoplan} focuses on goal-conditioned next-action planning. EgoThink~\citep{egothink} and VidEgoThink~\citep{videgothink} broaden first-person evaluation to perception, question answering, planning, grounding, and reward modeling. EgoTempo~\citep{egotempo} specifically targets fine-grained temporal understanding that requires integrating information across an egocentric video, whereas EgoProx~\citep{egoprox} evaluates embodied 3D proximity reasoning between the observer and surrounding objects. Although these benchmarks probe complementary aspects of egocentric intelligence, their prediction targets are primarily actions, semantic answers, spatial relations, or other symbolic outputs rather than the subsequent whole-frame visual state itself.

EgoExoBench~\citep{egoexobench} is the closest benchmark to EgoExo-Next. It provides a comprehensive evaluation of first- and third-person video understanding through semantic alignment, viewpoint association, and temporal reasoning. In particular, its temporal suite includes next-action prediction, action ordering, sequence alignment, and skill evaluation, while its cross-view tasks assess semantic and spatial correspondence between ego and exo observations. Thus, EgoExoBench already evaluates both temporal reasoning and cross-view understanding. The key distinction of EgoExo-Next is therefore not the presence of either capability individually, but the \emph{prediction target}: our next-state tasks require models to identify an unobserved subsequent visual state directly in the visual space.

This distinction is most explicit in \textbf{Ego-to-Exo Next-State}. Rather than predicting a subsequent action label or matching ego--exo content that is already observed, the model receives preceding egocentric observations and must identify the corresponding future exocentric visual state, $F_{1:t}^{ego} \rightarrow F_{t^{+}}^{exo}$. The task therefore composes temporal visual-state progression with viewpoint transformation within a single prediction. Moreover, the candidate set contains controlled mixtures of earlier-state and later-future distractors, preventing the target from being identified simply as the only future state or the most temporally advanced candidate. Consequently, EgoExo-Next complements existing egocentric and ego--exo benchmarks by directly evaluating \textbf{dynamic visual-state reasoning}, with Ego-to-Exo Next-State further testing its composition across time and viewpoint.

\section{Diagnosing Composed Ego-to-Exo Next-State Reasoning}
\label{sec:composed_diagnosis}

% --------Table4--------
\begin{table*}[t]
\centering
\caption{Diagnostic decomposition of ego-to-exo next-state reasoning on the matched subset. Results are accuracy (\%).}
\resizebox{\textwidth}{!}{
\begin{tabular}{lcccccc}
\toprule
\textbf{Setting}
& \textbf{GPT-5.2}
& \textbf{Gemini-3.1-Pro}
& \textbf{Qwen3-VL-235B}
& \textbf{Qwen3-VL-32B}
& \textbf{RoboBrain}
& \textbf{Avg.} \\
\midrule

Original Subtask 4
& 32.40 & 28.10 & 27.80 & 30.20 & 29.70 & 29.64 \\

+ Current Exo
& 48.70 & 44.00 & 45.10 & 38.30 & 40.10 & 43.24 \\

+ Next-State Ego
& 40.20 & 50.10 & 39.40 & 37.70 & 41.30 & 41.74 \\

+ Both
& 59.80 & 63.00 & 58.50 & 54.90 & 56.70 & 58.58 \\

\midrule

$\mathrm{S4}\mid\mathrm{S1,S2}_{E\rightarrow X},\mathrm{S3}$
& 60.50 & 59.20 & 57.40 & 54.10 & 55.80 & 57.40 \\

\bottomrule
\end{tabular}

}

\label{tab:reasoning_decomposition}
\end{table*}

We construct a matched subset of 200 Subtask 4 groups, with 100 Indoor and 100 Outdoor examples, and derive the corresponding Subtasks 1--3 from the same action clips and synchronized ego--exo views. We then introduce two oracle inputs: \textit{+ Current Exo} provides the synchronized exocentric frame $F_t^{\mathrm{exo}}$, reducing the cross-view correspondence requirement, while \textit{+ Next-State Ego} provides $F_{t^{+}}^{\mathrm{ego}}$, removing egocentric next-state prediction. \textit{+ Both} provides both inputs. We also report $\mathrm{S4}\mid\mathrm{S1,S2}_{E\rightarrow X},\mathrm{S3}$, the Subtask 4 accuracy when all three diagnostic tasks are answered correctly.

Table~\ref{tab:reasoning_decomposition} shows that the original Subtask 4 accuracy is 29.64\%.
(1) Cross-view correspondence and egocentric temporal progression both contribute to the task difficulty. Providing \textit{+ Current Exo} and \textit{+ Next-State Ego} increases accuracy to 43.24\% and 41.74\%, respectively.
(2) Integrating these component abilities remains challenging. Providing both oracle inputs raises accuracy to 58.58\%, while Subtask 4 accuracy reaches 57.40\% on groups where all three matched diagnostic tasks are answered correctly. These results support the proposed diagnostic decomposition, while the remaining errors indicate that combining the component abilities within a unified egocentric-to-exocentric next-state reasoning task remains difficult.

% ----------Effect of Caption Augmentation

\begin{figure}
    \centering
    \includegraphics[width=1\linewidth]{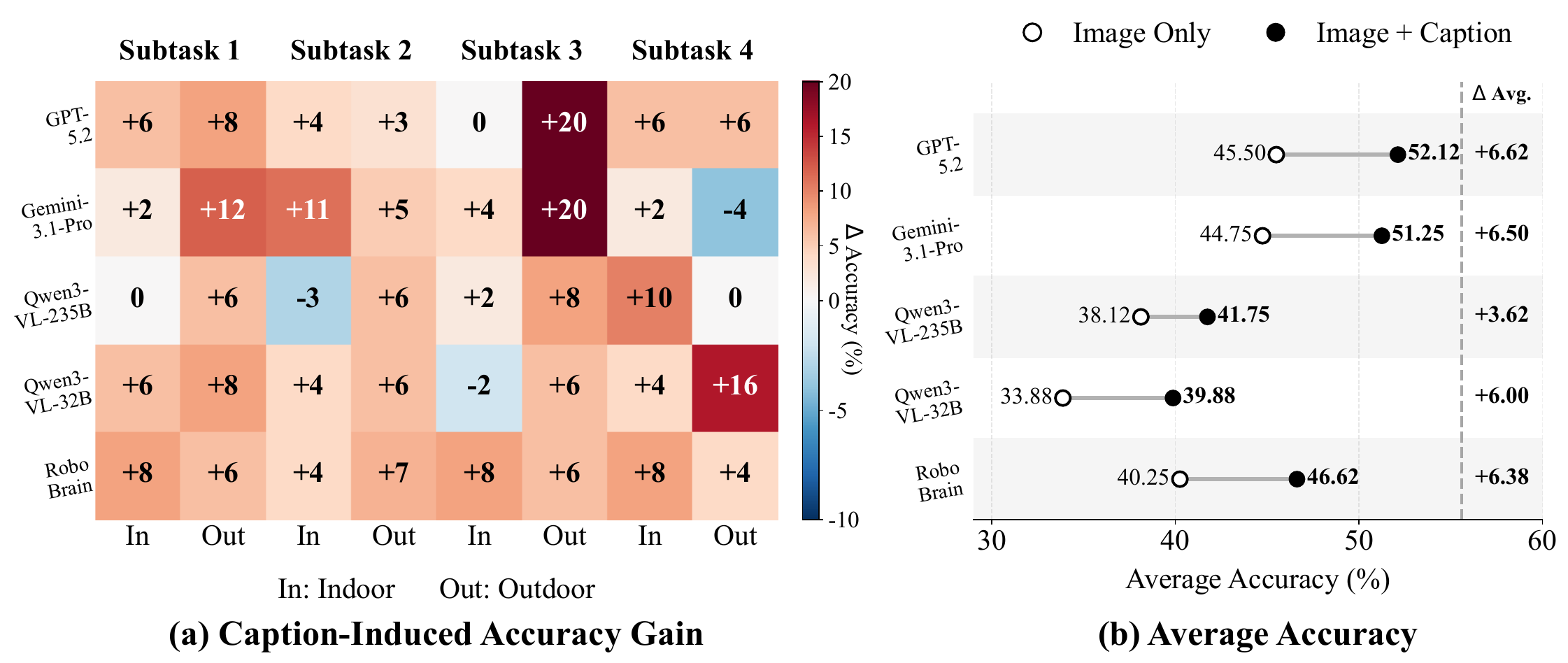}
    \caption{
    Effect of caption augmentation on EgoExo-Next. Left: accuracy gains across eight evaluation settings. Right: average accuracy under image-only and image-plus-caption inputs.
    }
    \label{fig:caption_augmented_results}
\end{figure}

\begin{figure}[t]
\centering
\begin{captionpromptbox}{Caption Template}
A \texttt{\textless Perspective\textgreater} view of \texttt{\textless Scene\textgreater} in a \texttt{\textless Setting\textgreater}. The image shows \texttt{\textless Situation\textgreater} involving \texttt{\textless Objects\textgreater}, with \texttt{\textless Background\textgreater} visible in the scene.
\end{captionpromptbox}

\caption{Prompt template used by Qwen2.5-VL-72B to generate captions for candidate images.}
\label{fig:CaptionAugmentation}
\end{figure}

\section{Effect of Caption Augmentation}\label{sec:caption_augmentation}

We investigate whether explicit textual descriptions can provide complementary semantic cues for visual-state reasoning. We sample 400 questions, with 50 examples from each Indoor and Outdoor split of the four subtasks, and evaluate models under two settings: \textit{Image Only} and \textit{Image + Caption}.

For each question, Qwen2.5-VL-72B independently generates a concise caption for each of the four candidate images using the template shown in Fig.~\ref{fig:CaptionAugmentation}. No captions are generated for the observation images, and Qwen2.5-VL-72B is excluded from the evaluated models. Human reviewers verify that each caption faithfully describes the visible perspective, scene, action or situation, objects, and background without introducing unsupported information; inaccurate descriptions are manually corrected. In the \textit{Image + Caption} setting, each caption is placed immediately after its corresponding candidate image and option label, while the observation inputs, candidate images, and evaluation prompts remain otherwise unchanged.

As shown in Fig.~\ref{fig:caption_augmented_results}, caption augmentation increases the overall accuracy from 40.50\% to 46.33\%, corresponding to a gain of 5.83 percentage points. Across 40 model--setting pairs, performance improves in 34 cases, remains unchanged in three, and decreases slightly in three. The largest average gain is observed on Subtask 3 Outdoor, where accuracy improves by 12.00 percentage points.

These results indicate that explicit semantic descriptions can complement visual information and help models distinguish candidate states. However, since the captions are generated directly from the candidate images, the improvement should be interpreted as evidence for the benefit of explicit semantic cues rather than as a direct measure of language bias.

\section{Additional Implementation Details and Results}
\label{sec:additional_details}

\subsection{Model and Inference Configurations}
\label{sec:model_configurations}

The proprietary models and Qwen3-VL-235B are evaluated through their official APIs, while the remaining open-source models are deployed from their official checkpoints on eight NVIDIA GeForce RTX 4090 GPUs. All evaluations are conducted in the zero-shot setting without in-context examples, task-specific fine-tuning, or gradient updates. For API-based evaluation, the temperature is set to 0, and the maximum response length is limited to 32 tokens. Each model is evaluated in three independent runs, and the mean accuracy is reported. The same evaluation instructions and output constraints are used across models.

\subsection{Human Evaluation Details}
\label{sec:human_evaluation_details}

Human performance is evaluated by three adult participants who were not involved in benchmark annotation or quality control. Each participant independently completes all 2,503 questions using the same observation and candidate images provided to the evaluated models. Candidate order is randomized, and participants are given access to all input images without a strict time limit. Table~\ref{tab:human_details} reports the participant-level accuracy for each Indoor and Outdoor split. Human performance in each setting is computed as the mean of the three individual accuracies. Following the main results, \textit{Avg.} denotes the unweighted mean over the eight subtask scene settings.

\begin{table*}[t]
\centering
\small
\setlength{\tabcolsep}{4.2pt}
\renewcommand{\arraystretch}{1.15}

\caption{Participant-level human performance on EgoExo-Next.
Each entry is computed from the participant's number
of correct answers in the corresponding evaluation split.}
\label{tab:human_details}

\begin{tabular}{lccccccccc}
\toprule
\multirow{2}{*}{Participant}
& \multicolumn{2}{c}{Subtask 1}
& \multicolumn{2}{c}{Subtask 2}
& \multicolumn{2}{c}{Subtask 3}
& \multicolumn{2}{c}{Subtask 4}
& \multirow{2}{*}{Avg.} \\
\cmidrule(lr){2-3}
\cmidrule(lr){4-5}
\cmidrule(lr){6-7}
\cmidrule(lr){8-9}

& Indoor & Outdoor
& Indoor & Outdoor
& Indoor & Outdoor
& Indoor & Outdoor
& \\

\midrule
Participant 1
& 97.05 & 99.66
& 99.67 & 99.34
& 99.71 & 98.42
& 96.12 & 97.48
& 98.43 \\

Participant 2
& 99.67 & 98.34
& 98.68 & 100.00
& 97.99 & 99.68
& 98.06 & 96.23
& 98.58 \\

Participant 3
& 98.36 & 99.41
& 99.67 & 99.67
& 98.85 & 99.05
& 97.09 & 96.86
& 98.62 \\

\midrule
Mean
& 98.36 & 99.14
& 99.34 & 99.67
& 98.85 & 99.05
& 97.09 & 96.86
& 98.55 \\

\bottomrule
\end{tabular}
\end{table*}

\end{document}